\documentclass[conference]{IEEEtran}
\IEEEoverridecommandlockouts
\usepackage{cite}
\usepackage{amsmath,amssymb,amsfonts}
\usepackage{algorithmic}
\usepackage{graphicx}
\usepackage{textcomp}
\usepackage[hidelinks]{hyperref}
\usepackage{subcaption}
\usepackage{xcolor}
\def\BibTeX{{\rm B\kern-.05em{\sc i\kern-.025em b}\kern-.08em
    T\kern-.1667em\lower.7ex\hbox{E}\kern-.125emX}}
\begin{document}
\bstctlcite{MyBSTcontrol}

\title{Crowd Counting on Heavily Compressed Images with Curriculum Pre-Training}

\author{Arian~Bakhtiarnia,
        Qi~Zhang,
        and~Alexandros~Iosifidis
\thanks{Arian Bakhtiarnia, Qi Zhang and Alexandros Iosifidis are with DIGIT, the Department of Electrical and Computer Engineering, Aarhus University, Aarhus, Midtjylland, Denmark (e-mail: arianbakh@ece.au.dk; qz@ece.au.dk; ai@ece.au.dk).}
\thanks{This work was funded by the European Union’s Horizon 2020 research and innovation programme under grant agreement No 957337, and by the Danish Council for Independent Research under Grant No. 9131-00119B.
}}

\maketitle

\begin{abstract}
JPEG image compression algorithm is a widely used technique for image size reduction in edge and cloud computing settings. However, applying such lossy compression on images processed by deep neural networks can lead to significant accuracy degradation. Inspired by the curriculum learning paradigm, we propose a training approach called curriculum pre-training (CPT) for crowd counting on compressed images, which alleviates the drop in accuracy resulting from lossy compression. We verify the effectiveness of our approach by extensive experiments on three crowd counting datasets, two crowd counting DNN models and various levels of compression. The proposed training method is not overly sensitive to hyper-parameters, and reduces the error, particularly for heavily compressed images, by up to 19.70\%.
\end{abstract}

\section{Introduction}

Many applications in smart cities, such as crowd monitoring, traffic surveillance and anomaly detection, utilize deep learning to process visual information \cite{bajovic2021marvel, https://doi.org/10.1002/itl2.187}. In such settings, typically the video frames taken by many cameras installed throughout the city are transmitted to a few edge or cloud servers to be processed. Since the capture resolution of modern cameras are typically high, in cases surpassing Full HD (1920$ \times $1080 pixels), transmitting raw images and video frames over the network results in massive bandwidth consumption and traffic congestion. JPEG compression is a common method used for reducing the size of images for transmission and storage. One of the benefits of JPEG is that it is readily available and configurable on many cameras. Moreover, JPEG encoding does not require a lot of computational power, which is crucial since capture devices are typically very limited in terms of computational resources. JPEG compression has been shown to provide a better accuracy-bandwidth trade-off compared to other simple compression techniques such as uniform downsampling and grayscaling \cite{bakhtiarnia2022analysis}. Other than reducing bandwidth, using heavily compressed images can be a computationally cheap approach to preserve privacy, since facial features will not be easily detectable, as shown in Figure \ref{fig:sample_patch} (f).
\begin{figure}[htbp]
\centerline{\includegraphics[width=0.5\textwidth]{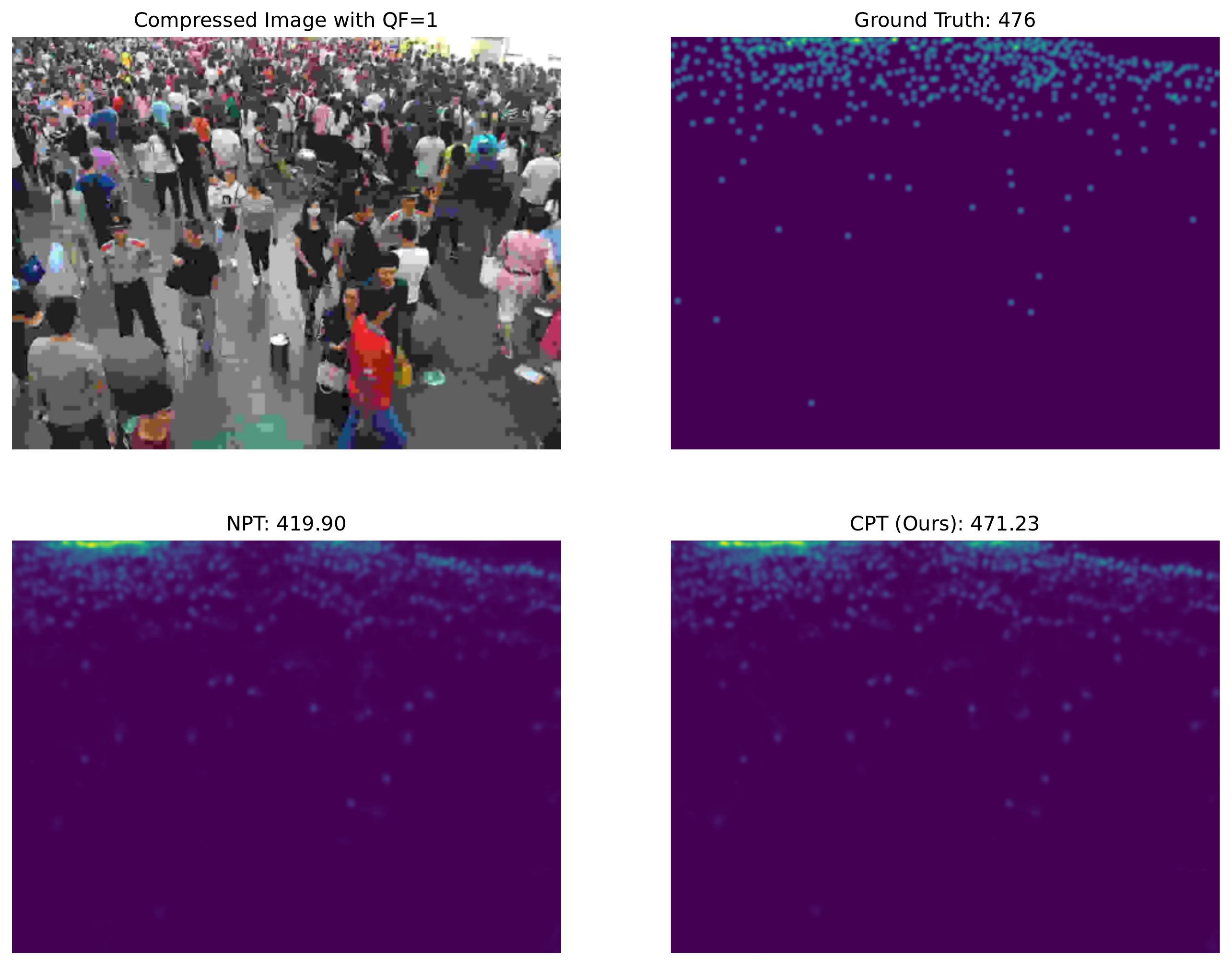}}
\caption{The proposed method (CPT) can achieve better performance compared to the normal pre-training (NPT) procedure. Even though the input image is heavily compressed with the lowest JPEG quality setting (i.e., quality factor $ \text{QF} = 1 $), our method can obtain a count close to ground truth. The 1024$ \times $768-pixel image is taken from the Shanghai Tech Part B dataset \cite{zhang2016single}.}
\label{fig:overview}
\end{figure}

JPEG is a lossy compression algorithm, meaning that the reconstructed image will not be exactly the same as the original image, since some visual information is lost during the encoding and decoding process. This negatively affects the performance of deep learning models, therefore, many methods exist that try to mitigate this loss of information. However, these methods typically introduce high overhead, are not optimized for particular downstream deep learning tasks, and do not focus on heavily compressed images. In this work, we propose a method called \textit{curriculum pre-training}, which alleviates the accuracy drop resulting from JPEG compression in the crowd counting task without introducing any overhead. Through extensive experiments, we show that the proposed method works well for both light and heavy compression in many situations, and is not overly sensitive to hyper-parameters. To the best of our knowledge, this is the first work that addresses the problem of crowd counting on heavily compressed images. Figure \ref{fig:overview} shows the result of the method applied to an example heavily compressed high-resolution image.

The rest of this paper is organized as follows. In the \textit{Related Work} section, we introduce relevant literature, including the role of JPEG compression in deep learning, crowd counting and curriculum learning. In the section titled \textit{Curriculum Pre-Training}, we describe our proposed method in detail. In the \textit{Experimental Setup and Results} section, we recount the setup and hyperparamers of our experiments, present the results and include ablation studies. Finally, section \textit{Conclusion} summarizes and discusses our results and provides directions for future research. Our code is publicly available\footnote{\url{https://gitlab.au.dk/maleci/curriculum-pre-training}}.

\section{Related Work}

\subsection{JPEG Compression in Deep Learning}

\begin{figure}
\begin{center}
\begin{tabular}{@{} c c c @{}}
\includegraphics[width=.14\textwidth]{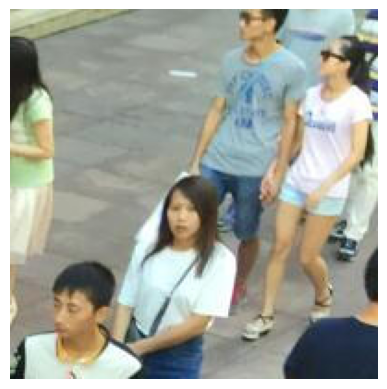}&
\includegraphics[width=.14\textwidth]{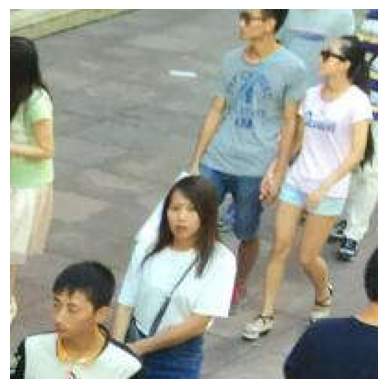}&
\includegraphics[width=.14\textwidth]{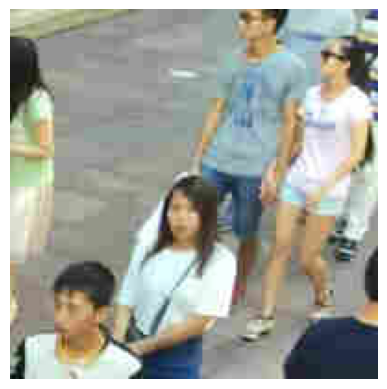}\\
(a) $ \text{QF}=75 $ & (b) $ \text{QF}=60 $ & (c) $ \text{QF}=15 $\\
\includegraphics[width=.14\textwidth]{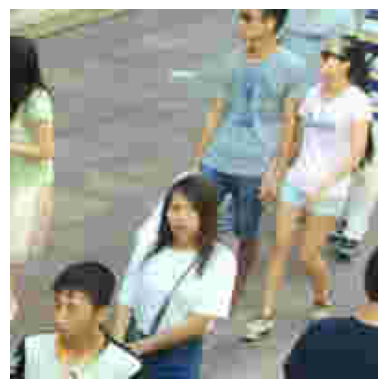}&
\includegraphics[width=.14\textwidth]{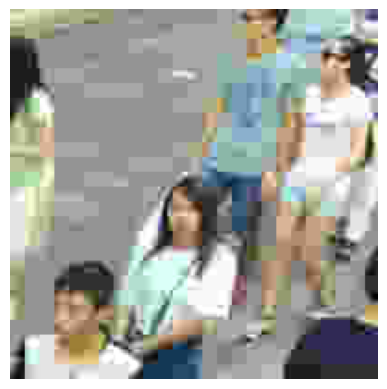}&
\includegraphics[width=.14\textwidth]{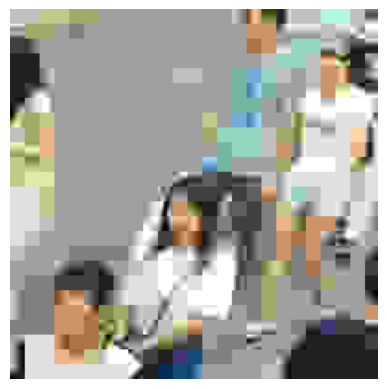}\\
(d) $ \text{QF}=10 $ & (e) $ \text{QF}=5 $ & (f) $ \text{QF}=1 $\\
\end{tabular}
\end{center}
\caption{Sample 200$ \times $200 pixel image patch taken from the Shanghai Tech Part B dataset \cite{zhang2016single}, after being compressed with a varying QF.}
\label{fig:sample_patch}
\end{figure}

JPEG is a lossy compression algorithm which can significantly reduce the size of images with minimal loss of visual information, and has built-in parameters for controlling the amount of compression. The JPEG encoding process consists of three main steps \cite{hudson2017jpeg, gueguen2018faster}. The first step is to convert the 24-bit 3-channel RGB image to the YCbCr color space based on the linear transformation:
\begin{multline}
\begin{small}
\begin{bmatrix}
Y\\
Cb\\
Cr
\end{bmatrix}
=
\begin{bmatrix}
0.299 & 0.587 & 0.114\\
-0.168935 & -0.331665 & 0.50059\\
0.499813 & -0.418531 & -0.081282
\end{bmatrix}
\begin{bmatrix}
R\\
G\\
B
\end{bmatrix}.
\label{eq:rgb2ycbcr}
\end{small}
\end{multline}
Since the human eye is less sensitive to color details represented by the chroma components Cb and Cr, rather than brightness details represented by the luma component Y, Cb and Cr are downsampled by a factor of 2 or 3. The second step is to split each component to 8$ \times $8 blocks and convert the information to frequency domain by performing a two-dimensional discrete cosine transform (DCT) on each block. The amplitude of the frequency domain information is then quantized based on
\begin{equation}
Q = D \oslash T_s,
\label{eq:quantize_hadamard}
\end{equation}
where $ D $ contains the obtained DCT amplitudes, $ \oslash $ is element-wise matrix division (Hadamard division), and $ T_s $ is derived based on
\begin{equation}
T_{s_{ij}} = \Big\lfloor \frac{s T_{b_{ij}} + 50}{100} \Big\rfloor,
\label{eq:quantize_ts}
\end{equation}
where
\begin{equation}
s = \begin{cases}
  \frac{5000}{\text{QF}}, & 1 \leq \text{QF} < 50, \\
  200 - 2q, & 50 \leq \text{QF} \leq 100,
\end{cases}
\label{eq:quantize_s}
\end{equation}
and $ T_b $ is the base quantization matrix:
\begin{equation}
T_b = \begin{bmatrix}
16 & 11 & 10 & 16 & 24 & 40 & 51 & 61\\
12 & 12 & 14 & 19 & 26 & 58 & 60 & 55\\
14 & 13 & 16 & 24 & 40 & 57 & 69 & 56\\
14 & 17 & 22 & 29 & 51 & 87 & 80 & 62\\
18 & 22 & 37 & 56 & 68 & 109 & 103 & 77\\
24 & 35 & 55 & 64 & 81 & 104 & 113 & 92\\
49 & 64 & 78 & 87 & 103 & 121 & 120 & 101\\
72 & 92 & 95 & 98 & 112 & 100 & 103 & 99
\end{bmatrix}.
\label{eq:quantize_tb}
\end{equation}
The quality setting $ 1 \leq QF \leq 100 $ in equation \ref{eq:quantize_s} is an integer number that controls the amount of quantization, with 1 corresponding to the lowest quality and 100 the highest. Figure \ref{fig:sample_patch} shows a sample image at different quality settings. In line with the findings of \cite{dodge2016understanding}, the quality of the image stays relatively high with $ \text{QF} \geq 10 $. The last step in the JPEG encoding process is to further reduce the size of the quantized matrices using Huffman encoding.

JPEG compression has been shown to reduce the performance of deep neural networks, particularly with high compression settings \cite{dodge2016understanding}. Therefore, there have been efforts to improve the quality of images compressed using JPEG  \cite{ehrlich2021analyzing}. However, JPEG artifact correction methods have several shortcomings. Some methods focus on improving the visual quality of the reconstructed images without paying attention to the downstream tasks \cite{jiang2021towards}. Even though such methods try to reconstruct the images as closely as possible to the original non-compressed ones, it is not clear whether such reconstructions lead to optimal performance in a particular deep learning task. Furthermore, JPEG artifact correction methods typically ignore heavily compressed images, defined as having a $ \text{QF} < 10 $, since they claim there is little information preserved below this threshold. However, as we show in this work, heavy compression can still offer valuable options in the trade-off between size and performance for the crowd counting task. Finally, modern JPEG artifact correction methods use deep neural networks to improve the quality of the reconstructed images, which adds high overhead to an already demanding task. In contrast, our method is designed for and evaluated on the downstream task, and it can perform well even under heavy compression without adding any overhead as it only modifies the training procedure of the task DNN (deep neural network).

\subsection{Crowd Counting}

Crowd counting is the task of counting the total number of people present in a given scene \cite{gao2020cnn}. The input images or video frames are typically of high-resolution, and the expected output is a number for each image corresponding to the count. However, crowd counting methods typically also output a density map detailing the density of the crowd at each location of the image. Crowd counting datasets provide head annotations as ground truth labels, which are the locations of the center of the head for each person in the image.

Crowd counting methods are usually evaluated based on mean absolute error (MAE) and mean squared error (MSE) which are measures for accuracy and robustness, respectively \cite{10.1145/3460426.3463628}. In this work, we evaluate the performance of DNNs based on MAE since accuracy is our primary goal. \cite{bakhtiarnia2022analysis} compares the accuracy of crowd counting methods on images compressed with different simple lossy compression algorithms, and shows that JPEG compression offers the best trade-off between accuracy and size.

\subsection{Curriculum Learning}

Curriculum learning is a training paradigm for deep neural networks, which is inspired by how humans learn in their formal education where a knowledgeable teacher starts the course with simple concepts and gradually increases the difficulty of the material. Similarly, the training examples for neural networks can be sorted based on some measure of difficulty. Training can start with the simplest examples and harder examples can be gradually introduced during the training process, which can ultimately lead to higher accuracy \cite{bengio2009curriculum}. For instance, in image classification, images with complex backgrounds may be more difficult for a DNN to classify. In this case, the confidence of another ``teacher'' DNN on each images can be used as a measure of difficulty.

In this way, curriculum learning is built by using two main functions: the sorting function that assigns a difficulty to each training example, and the pacing function that determines the pace for introducing harder examples in the training process. Various sorting and pacing functions have been explored in the literature for many deep learning tasks \cite{DBLP:journals/corr/abs-2010-13166}. Curriculum learning is very sensitive to the choice of scoring and pacing functions and their hyper-parameters \cite{pmlr-v97-hacohen19a}. It should be noted that as opposed to human learning, sometimes the opposite approach of starting the training from the hardest examples, called \textit{anti-curriculum}, works best for DNNs \cite{pmlr-v97-hacohen19a, bakhtiarnia2021improving}.

Curriculum learning has been explored for improving crowd counting accuracy, where a weight is assigned to each pixel in the density map \cite{9275392}. However, that method operates on high-quality images and the effect of image compression is not considered. In this work, we focus on how to mitigate the accuracy degradation resulting from heavy image compression, and not on improving crowd counting in high-quality images.

\begin{figure}[htbp]
\centerline{\includegraphics[width=0.4\textwidth]{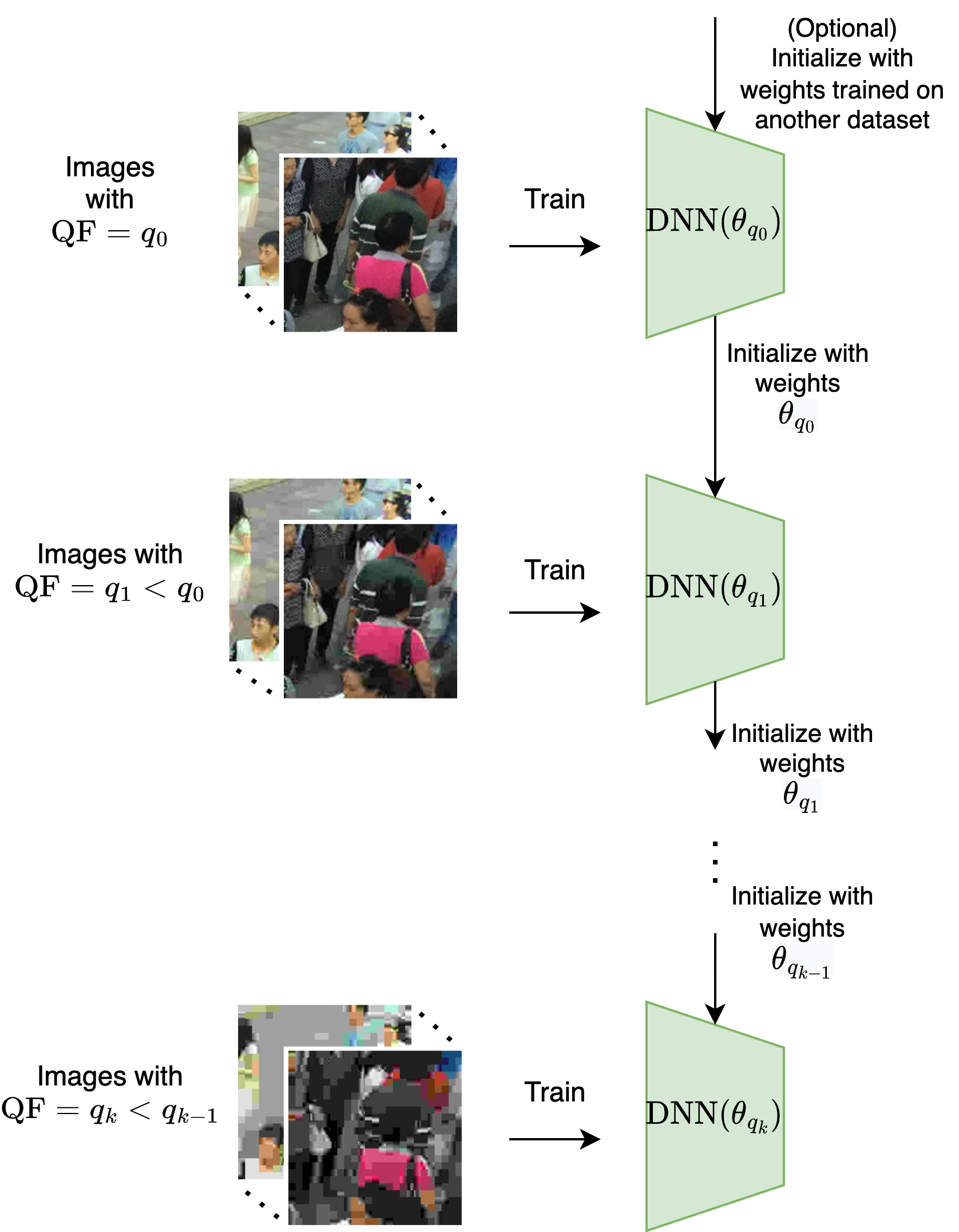}}
\caption{Curriculum pre-training (CPT) procedure.}
\label{fig:method}
\end{figure}

\section{Curriculum Pre-Training}

Given compressed images, the simplest approach would be to use a DNN trained on high-quality images to produce an output for the task. However, this leads to significant drops in accuracy. A more sensible approach would be to train the DNN on the compressed images. To further increase the accuracy, training can be initialized using pre-trained weights taken from the DNN trained on the original high-quality images, and then fine-tuned using compressed images. We call this approach \textit{normal pre-training (NPT)} and compare the proposed method to this baseline.

The quality setting (QF) in JPEG encoding can be viewed as a natural scoring function for curriculum learning, where high quality images with high QF can be viewed as ``easy'', and more heavily compressed images with low QF can be viewed as ``difficult''. However, in curriculum learning, the final accuracy of the network is evaluated on all examples, including easy and difficult ones. In contrast, in this task, our goal is to obtain optimal accuracy for a particular quality setting. Therefore, we only care about the accuracy of the most difficult examples. As a result, we need a special pacing function that removes the easier examples as the training progresses.

The proposed method called \textit{curriculum pre-training (CPT)} trains the neural network for successive lower QFs in a step-by-step manner, using the trained weights of the previous step to initialize the weights for the next step in a cascaded fashion. Assuming the quality setting of the original images in the training set is $ q_0 $, and our goal is to obtain the best accuracy for images of quality $ q_k < q_0 $, we define a curriculum $ C = (q_0, q_1, \dots, q_k) $ where $ q_i > q_j $ if $ i < j $. We start by training the DNN on images of quality $ q_0 $. After the training is finished, we use the trained weights $ \theta_{q_0} $ to initialize the DNN, and then train the DNN on images of quality $ q_1 $. We continue this process until we obtain $ \theta_{q_k} $. This procedure is illustrated in Figure \ref{fig:method}.

The intuition behind this approach is that optima for images of quality $ q_0 $ might be drastically different from optima for images of quality $ q_k $. Therefore, starting the training for images of quality $ q_k $ from an optimum for images of quality $ q_0 $ might lead to a convergence of the parameters of the network in an undesired location of the loss landscape. On the other hand, as with many deep learning tasks, starting with no pre-trained weights can lead to sub-optimal results. Since the optima for similar quality settings are more likely to be close to each other, by gradually shifting the starting location for successive image qualities, we can reap the benefits of pre-training with lower risk of the final parameters falling in an undesired location of the loss landscape.

There are some crucial differences between the proposed method and the typical curriculum learning. First, in the proposed method, each iteration contains only images of a particular difficulty, whereas in typical curriculum learning there is a mixture of difficulties in each iteration. Second, the pacing of curriculum learning is usually much faster, and the most difficult examples are introduced after only a handful of epochs \cite{pmlr-v97-hacohen19a, bakhtiarnia2021improving}, whereas in the proposed method the DNN is trained on each difficulty for many epochs. Finally, we reset the training hyper-parameters (including learning rate and weight decay) before moving on to the next quality setting.

\section{Experiments}

\subsection{Hyper-Parameters and Setup}

In this work, we use the Shanghai Tech \cite{zhang2016single} and DISCO \cite{hu2020ambient} datasets. Shanghai Tech is a widely used crowd counting dataset which consists of two parts: part A contains 482 images of very dense crowds taken from the web with variable sizes ranging from 420$ \times $182 pixels to 1,024$ \times $1,024 pixels, and part B contains 716 images of moderate density taken from a busy street with size 1,024$ \times $768 pixels. Both parts split the images into training and test sets. Since these datasets do not provide validation data, we randomly take 20\% from the training set of part A and 10\% from part B as validation data\footnote{The random seed and selection procedure can be found in our source code.}. SASNet \cite{song2021choose} is the state-of-the-art DNN for crowd counting on the Shanghai Tech dataset at the time of this writing. SASNet uses the first 10 layers of VGG16 \cite{simonyan2014very} as a feature extractor, and adds many other layers on top of this architecture in order to extract and combine features across multiple scales.

DISCO is a challenging crowd counting dataset taken from a large variety of scenes which include diverse illumination settings such as day and night. DISCO contains 1,935 images split into training, validation and test sets. 
CSRNet \cite{li2018csrnet} is a high-performing DNN for crowd counting on the DISCO dataset. Similar to SASNet, CSRNet also uses the first 10 layers of VGG16 as a feature extractor and adds 6 dilated convolution layers after. The images in all of the aforementioned datasets are saved in the JPEG format and are already compressed with a quality setting of 75.

In our experiments, we choose the curriculum $ C = (75, 60, 40, 30, 25, 15, 10, 5, 1) $ for training on images with $ \text{QF} = 1 $, and to train on images with a higher QF we use the subset of this curriculum down to (and including) that particular QF. For instance, for training on images with $ \text{QF} = 25 $ we use the curriculum $ C' = (75, 60, 40, 30, 25) $. These quality settings are chosen such that the difference between the average size of the images of successive quality settings are roughly the same, and relatively small. Table \ref{tab:hyperparameters} shows the hyper-parameter values and setup for each set of experiments. The best learning rate is chosen for each set of experiments from $ \text{LR} = \{10^{-3}, 10^{-4}, \dots, 10^{-7}\} $. The training procedure for CSRNet trained on DISCO is similar to \cite{hu2020ambient}. As previously mentioned, the Shanghai Tech Part A dataset has images of variable size, therefore, the batch size needs to be 1 since PyTorch \cite{paszke2019pytorch} only allows training on batches of images with the same size and training was done on only a single GPU. All experiments were repeated twice and the average error and standard deviation were recorded.

\begin{table}[htbp]
\caption{Hyper-parameter values and setup of the experiments.}
\begin{center}
\resizebox{\linewidth}{!}{
\begin{tabular}{ c c | c c c c c c c } 
\hline

Dataset & DNN & Optimizer & LR$^*$ & LRD$^\|$ & WD$^\dagger$ & Epochs & BS$^{**}$ & Hardware \\

\hline
\hline

DISCO & CSRNet & AdamW$^\ddagger$ & $ 10^{-5} $ & 0.99 & $ 10^{-4} $ & 100 & 16 & 3$ \times $Nvidia A6000\\
SHTB$^\mathsection$ & SASNet & AdamW & $ 10^{-5} $ & 0.99 & $ 10^{-4} $ & 100 & 5 & 3$ \times $Nvidia A6000\\
SHTA$^\mathparagraph$ & SASNet & AdamW & $ 10^{-7} $ & 0.99 & $ 10^{-6} $ & 50 & 1 & 1$ \times $Nvidia A6000\\

\hline
\multicolumn{9}{l}{$^*$Learning rate}\\
\multicolumn{9}{l}{$^\|$Learning rate decay per epoch}\\
\multicolumn{9}{l}{$^{**}$Batch size per GPU}\\
\multicolumn{9}{l}{$^\dagger$Weight decay}\\
\multicolumn{9}{l}{$^\ddagger$\cite{loshchilov2018decoupled}}\\
\multicolumn{9}{l}{$^\mathsection$Shanghai Tech Part B}\\
\multicolumn{9}{l}{$^\mathparagraph$Shanghai Tech Part A}\\
\end{tabular}
}
\end{center}
\label{tab:hyperparameters}
\end{table}

\subsection{Results}

\begin{figure}[htbp]
\centerline{\includegraphics[width=0.44\textwidth]{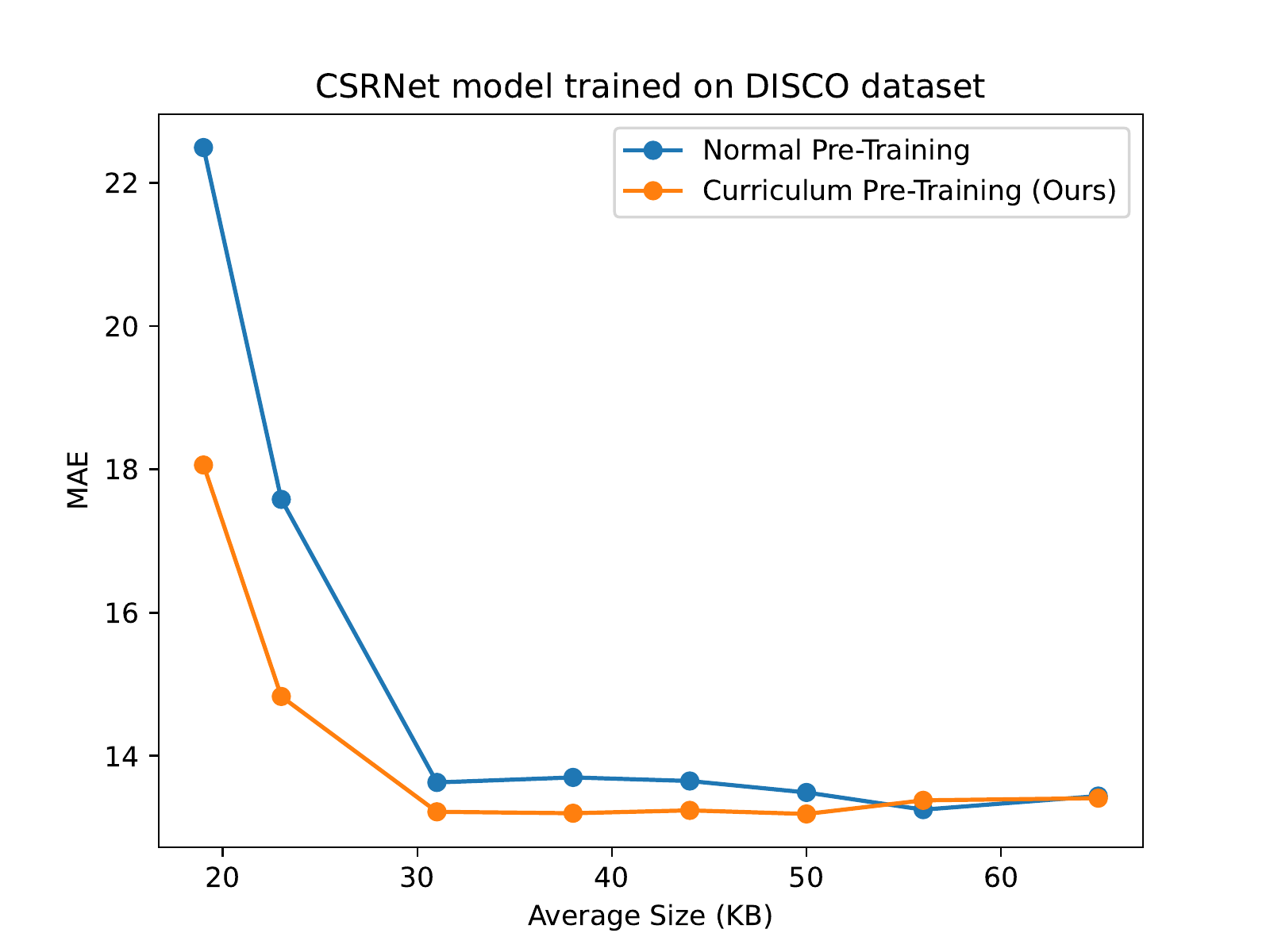}}
\caption{Trade-off between accuracy and image size with NPT and CPT using CSRNet \cite{li2018csrnet} on DISCO dataset \cite{hu2020ambient}.}
\label{fig:csrnet_disco}
\end{figure}

\begin{figure}[htbp]
\centerline{\includegraphics[width=0.44\textwidth]{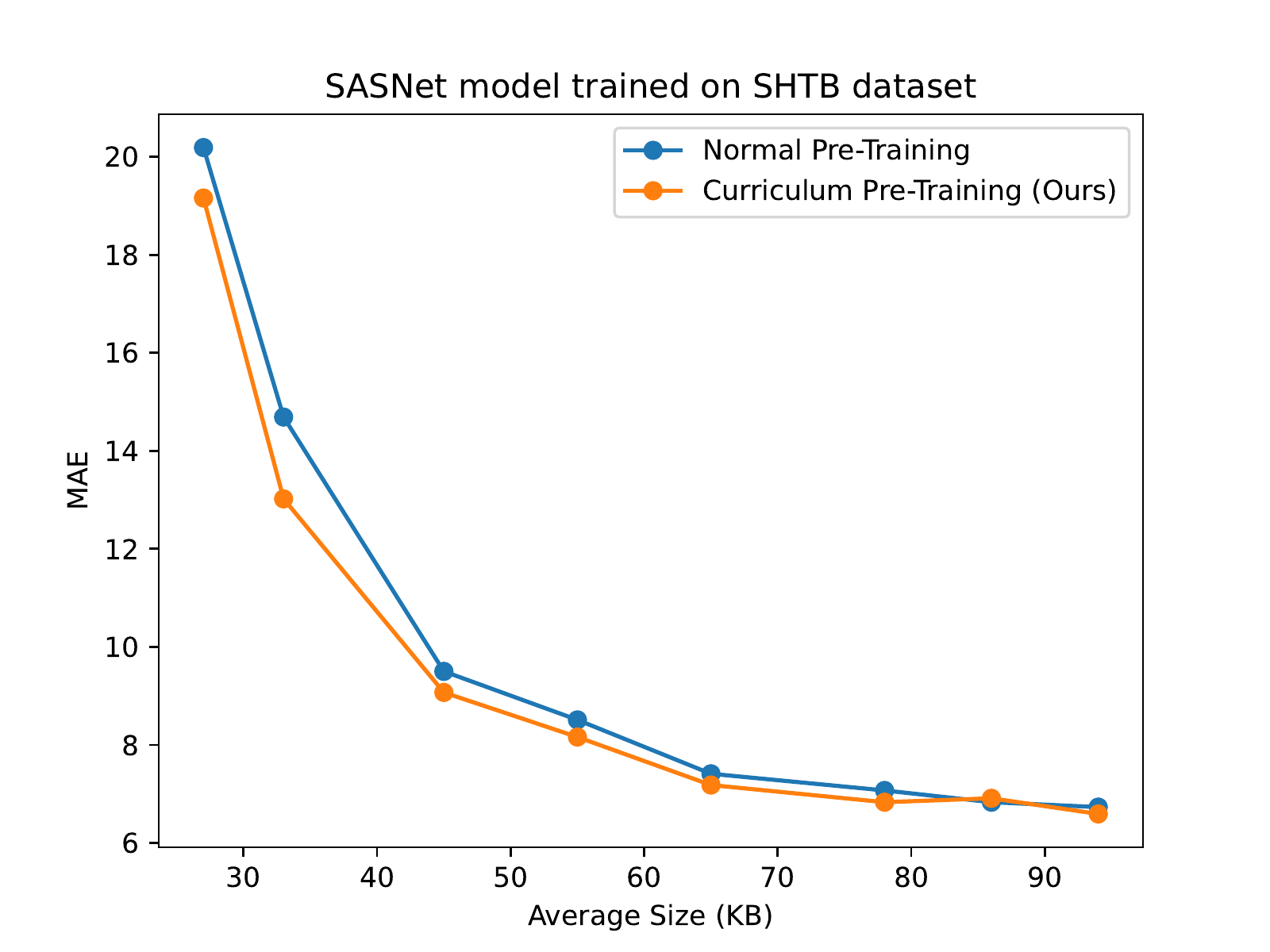}}
\caption{Trade-off between accuracy and image size with NPT and CPT using SASNet \cite{song2021choose} on Shanghai Tech Part B dataset \cite{zhang2016single}.}
\label{fig:sasnet_shtb}
\end{figure}

\begin{figure}[htbp]
\centerline{\includegraphics[width=0.44\textwidth]{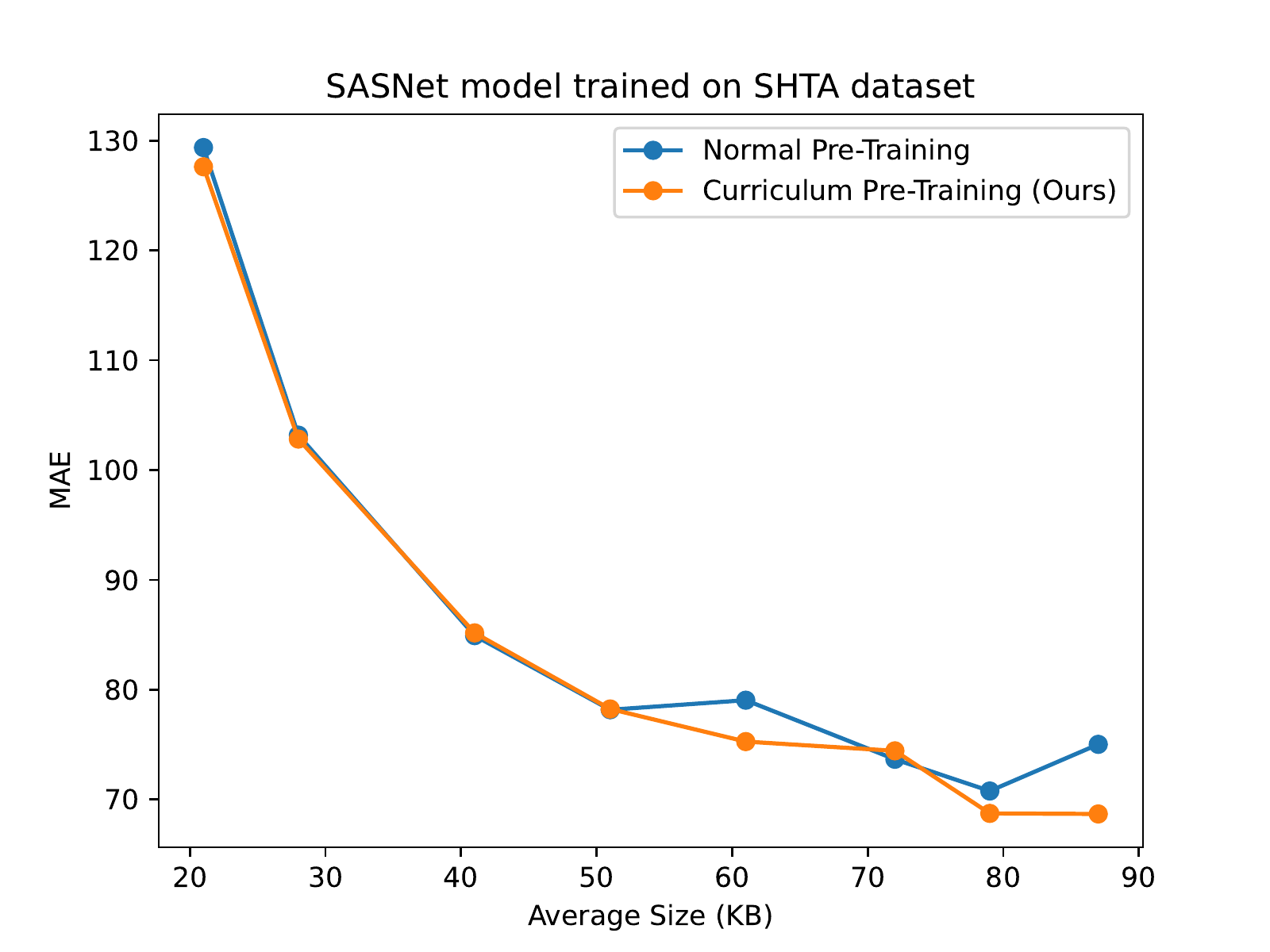}}
\caption{Trade-off between accuracy and image size with NPT and CPT using SASNet \cite{song2021choose} on Shanghai Tech Part A dataset \cite{zhang2016single}.}
\label{fig:sasnet_shta}
\end{figure}

The results for CSRNet on DISCO, SASNet on Shanghai Tech Part B and SASNet on Shanghai Tech Part A are shown in Tables \ref{tab:disco}, \ref{tab:shtb} and \ref{tab:shta}, respectively, and in Figures \ref{fig:csrnet_disco}, \ref{fig:sasnet_shtb} and \ref{fig:sasnet_shta}, respectively. In the Tables, the lowest error is highlighted for QF value.

\begin{figure}
\begin{center}
\begin{tabular}{ c c }
\includegraphics[width=.2\textwidth]{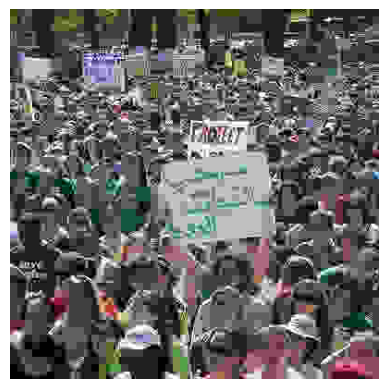}&
\includegraphics[width=.2\textwidth]{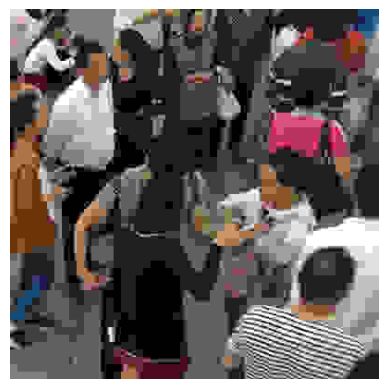}\\
(a) Shanghai Tech Part A & (b) Shanghai Tech Part B\\
\end{tabular}
\end{center}
\caption{Sample 400$ \times $400 pixel image patch taken from the Shanghai Tech Part A and Part B datasets \cite{zhang2016single} with $ \text{QF}=1 $. While features such as outline and clothing are still visible under heavy compression in sparsely crowded scenes such as (b), heavy compression may lead to excessive loss of visual information in densely crowded scenes such as (a).}
\label{fig:info_loss}
\end{figure}

It can be observed that curriculum pre-training achieves a higher accuracy in 19 out of 24 cases. Generally, the heavier the compression gets, the higher the improvement obtained by curriculum pre-training is, compared to normal pre-training. The only exceptions are some of the experiments on Shanghai Tech Part A, perhaps because the loss of information has much greater impact in very densely crowded scenes where only parts of head are visible, as shown in Figure \ref{fig:info_loss}. However, even on Shanghai Tech Part A, curriculum pre-training obtains the best performance for the heaviest compression.

In addition, it is known that JPEG compression can sometimes benefit the accuracy due to increased contrast between the foreground and background, which happens as a result of unequal quantization performed by JPEG on different DCT coefficients. Because quantization is non-linear, it reduces more energy in the background than the foreground \cite{yang2021compression}. This effect is also visible in some of our experiments. For instance, in Table \ref{tab:disco}, using images compressed with a $ \text{QF} = 25 $ leads to a lower error compared to using $ \text{QF} = 40 $.

\begin{table}[htbp]
\caption{Performance of curriculum pre-training of CSRNet \cite{li2018csrnet} on DISCO dataset \cite{hu2020ambient}. Lowest error for each QF is highlighted.}
\begin{center}
\resizebox{\linewidth}{!}{
\begin{tabular}{ c c c c c } 
\hline

JPEG QF & Avg. Size & NPT$^{*}$ MAE & CPT$^{\dagger}$ MAE (Ours) & Improvement\\

\hline
\hline

75 & 113 KB & 13.23 ± 0.08 & - & -\\
60 & 86 KB & 13.11 ± 0.13 & - & -\\
\hline
40 & 65 KB & 13.44 ± 0.21 & \textbf{13.41 ± 0.38} & 0.22\% \\
30 & 56 KB & \textbf{13.25 ± 0.01} & 13.38 ± 0.07 & -0.98\% \\
25 & 50 KB & 13.49 ± 0.08 & \textbf{13.19 ± 0.29} & 2.22\% \\
20 & 44 KB & 13.65 ± 0.19 & \textbf{13.24 ± 0.27} & 3.00\% \\
15 & 38 KB & 13.70 ± 0.08 & \textbf{13.20 ± 0.13} & 3.65\% \\
10 & 31 KB & 13.63 ± 0.10 & \textbf{13.22 ± 0.17} & 3.01\% \\
5 & 23 KB & 17.58 ± 0.07 & \textbf{14.83 ± 0.36} & 15.64\% \\
1 & 19 KB & 22.49 ± 0.11 & \textbf{18.06 ± 0.06} & 19.70\% \\

\hline
\multicolumn{4}{l}{$^{*}$Normal Pre-Training}\\
\multicolumn{4}{l}{$^{\dagger}$Curriculum Pre-Training}\\
\end{tabular}
}
\end{center}
\label{tab:disco}
\end{table}

\begin{table}[htbp]
\caption{Performance of curriculum pre-training of SASNet \cite{song2021choose} on Shanghai Tech Part B dataset \cite{zhang2016single}. Lowest error for each QF is highlighted.}
\begin{center}
\resizebox{\linewidth}{!}{
\begin{tabular}{ c c c c c c } 
\hline

JPEG QF & Avg. Size & NPT$^{*}$ MAE & CPT$^{\dagger}$ MAE (Ours) & Improvement\\

\hline
\hline

75 & 168 KB & 6.31$^{\ddagger}$ & - & -\\
60 & 147 KB & 6.64 ± 0.06 & - & -\\
\hline
40 & 94 KB & 6.73 ± 0.04 & \textbf{6.59 ± 0.06} & 2.10\%\\
30 & 86 KB & \textbf{6.83 ± 0.01} & 6.91 ± 0.04 & -1.17\%\\
25 & 78 KB & 7.07 ± 0.00 & \textbf{6.83 ± 0.06} & 3.39\%\\
20 & 65 KB & 7.41 ± 0.00 & \textbf{7.18 ± 0.07} & 3.10\%\\
15 & 55 KB & 8.51 ± 0.13 & \textbf{8.16 ± 0.06} & 4.11\%\\
10 & 45 KB & 9.50 ± 0.14 & \textbf{9.07 ± 0.09} & 4.53\%\\
5 & 33 KB & 14.69 ± 0.10 & \textbf{13.02 ± 0.06} & 11.37\%\\
1 & 27 KB & 20.19 ± 0.18 & \textbf{19.16 ± 0.09} & 5.10\%\\

\hline
\multicolumn{4}{l}{$^{*}$Normal Pre-Training}\\
\multicolumn{4}{l}{$^{\dagger}$Curriculum Pre-Training}\\
\multicolumn{4}{l}{$^{\ddagger}$Pre-trained Weights from SASNet \cite{song2021choose}, thus not repeated}\\
\end{tabular}
}
\end{center}
\label{tab:shtb}
\end{table}

\begin{table}[htbp]
\caption{Performance of curriculum pre-training of SASNet \cite{song2021choose} on Shanghai Tech Part A dataset \cite{zhang2016single}. Lowest error for each QF is highlighted.}
\begin{center}
\resizebox{\linewidth}{!}{
\begin{tabular}{ c c c c c c } 
\hline

JPEG QF & Avg. Size & NPT$^{*}$ MAE & CPT$^{\dagger}$ MAE (Ours) & Improvement\\

\hline
\hline

75 & 150 KB & 54.12$^{\ddagger}$ & - & -\\
60 & 129 KB & 67.31/70.11 & - & -\\
\hline
40 & 87 KB & 75.02 ± 6.20 & \textbf{68.68 ± 4.64} & 8.45\%\\
30 & 79 KB & 70.77 ± 3.06 & \textbf{68.73 ± 0.23} & 2.88\%\\
25 & 72 KB & \textbf{73.66 ± 1.97} & 74.43 ± 3.59 & -1.05\%\\
20 & 61 KB & 79.04 ± 7.25 & \textbf{75.27 ± 0.42} & 4.77\%\\
15 & 51 KB & \textbf{78.17 ± 1.67} & 78.24 ± 5.71 & -0.01\%\\
10 & 41 KB & \textbf{84.92 ± 1.22} & 85.17 ± 3.85 & -0.03\%\\
5 & 28 KB & 103.18 ± 4.43 & \textbf{102.82 ± 5.16} & 0.03\%\\
1 & 21 KB & 129.37 ± 1.38 & \textbf{127.62 ± 4.64} & 1.35\%\\

\hline
\multicolumn{4}{l}{$^{*}$Normal Pre-Training}\\
\multicolumn{4}{l}{$^{\dagger}$Curriculum Pre-Training}\\
\multicolumn{4}{l}{$^{\ddagger}$Pre-trained Weights from SASNet \cite{song2021choose}, thus not repeated}\\
\end{tabular}
}
\end{center}
\label{tab:shta}
\end{table}

\subsection{Ablation Studies}

Table \ref{tab:ablation_shtb} shows the results of ablation studies for CSRNet \cite{li2018csrnet} on the DISCO dataset \cite{hu2020ambient} with $ \text{QF}=1 $. From the first row, it can be observed that using weights from $ \text{QF}=75 $ directly for inference on images with $ \text{QF}=1 $, without any fine-tuning, leads to a very high error rate. Furthermore, the second and third rows show that pre-trained weights from $ \text{QF}=75 $ do not help compared to training from scratch. In fact, they might even lead to slightly higher error, although this is usually not the case. For instance, with $ \text{QF}=5 $, pre-trained weights from $ \text{QF}=75 $ lead to an MAE of 17.58 ± 0.07, compared to an MAE of 18.13 ± 0.32 obtained by training from scratch. In addition, the fourth row shows that just using the pre-trained weights from a close quality setting does not result in the best accuracy, and the cascaded nature of training is valuable. Moreover, it can be seen in the last three rows that the proposed method is resilient to both variable number of steps in the curriculum as well as to variations in the quality settings in the curriculum.

\begin{table}[htbp]
\caption{Ablation studies for CSRNet \cite{li2018csrnet} on DISCO dataset \cite{hu2020ambient} with $ \text{QF}=1 $.}
\begin{center}
\resizebox{\linewidth}{!}{
\begin{tabular}{ c l c } 
\hline

Row \# & Training Method & MAE\\

\hline
\hline

1 & No fine-tuning with weights from $ \text{QF}=75 $ & 84.00 ± 4.29 \\
2 & No pre-training$^\ddagger$ & 22.05 ± 0.16 \\
3 & Pre-trained weights from $ \text{QF}=75^{*} $ & 22.49 ± 0.11\\
4 & Pre-trained weights from $ \text{QF}=5 $ & 20.24 ± 1.39 \\
5 & Curriculum pre-training with $ C=(75, 60, 40, 30, 25, 20, 15, 10, 5)^{\dagger} $ & 18.06 ± 0.06\\
6 & Curriculum pre-training with $ C=(75, 40, 25, 15, 5) $ & 18.01 ± 0.13 \\
7 & Curriculum pre-training with $ C=(75, 45, 28, 17, 6) $ & 18.23 ± 0.31 \\

\hline
\multicolumn{3}{l}{$^\ddagger$In PyTorch, weights are initialized from a uniform distribution by default \cite{paszke2019pytorch})}\\
\multicolumn{3}{l}{$^{*}$Equivalent of normal pre-training presented in Table \ref{tab:disco}}\\
\multicolumn{3}{l}{$^{\dagger}$Equivalent of curriculum pre-training presented in Table \ref{tab:disco}}\\
\end{tabular}
}
\end{center}
\label{tab:ablation_shtb}
\end{table}

\section{Conclusion}

We showed that the proposed curriculum pre-training method can improve the accuracy of crowd counting DNNs that process compressed images. Since the method only modifies the weights of the DNN during the training phase, it does not add any overhead to the overall task. This is crucial as crowd counting is already a very demanding task since crowd counting methods need to process high-resolution inputs. Moreover, we showed that the proposed method works particularly well for heavily compressed images. In the provided ablation studies, we showed that the method is not overly sensitive to hyper-parameters, and that slight variations of hyper-parameters lead to similar results.

Even though we focused on crowd counting in this work, no part of the proposed method depends on the particular crowd counting setting. Therefore, it is reasonable to assume that this method can be used to improve the accuracy of DNNs processing compressed images in other deep learning tasks, particularly for other dense regression problems such as depth estimation, and other deep learning applications used in smart cities, for instance, crowd anomaly detection.

\bibliographystyle{IEEEtran}
\bibliography{references.bib}

\begin{thebibliography}{10}
\providecommand{\url}[1]{#1}
\csname url@samestyle\endcsname
\providecommand{\newblock}{\relax}
\providecommand{\bibinfo}[2]{#2}
\providecommand{\BIBentrySTDinterwordspacing}{\spaceskip=0pt\relax}
\providecommand{\BIBentryALTinterwordstretchfactor}{4}
\providecommand{\BIBentryALTinterwordspacing}{\spaceskip=\fontdimen2\font plus
\BIBentryALTinterwordstretchfactor\fontdimen3\font minus
  \fontdimen4\font\relax}
\providecommand{\BIBforeignlanguage}[2]{{%
\expandafter\ifx\csname l@#1\endcsname\relax
\typeout{** WARNING: IEEEtran.bst: No hyphenation pattern has been}%
\typeout{** loaded for the language `#1'. Using the pattern for}%
\typeout{** the default language instead.}%
\else
\language=\csname l@#1\endcsname
\fi
#2}}
\providecommand{\BIBdecl}{\relax}
\BIBdecl

\bibitem{bajovic2021marvel}
D.~Bajovic, A.~Bakhtiarnia \emph{et~al.}, ``Marvel: Multimodal extreme scale
  data analytics for smart cities environments,'' \emph{BalkanCom}, 2021.

\bibitem{https://doi.org/10.1002/itl2.187}
S.~Bhattacharya, S.~R.~K. Somayaji \emph{et~al.}, ``A review on deep learning
  for future smart cities,'' \emph{Internet Technology Letters}, 2022.

\bibitem{bakhtiarnia2022analysis}
A.~Bakhtiarnia, B.~Leporowski \emph{et~al.}, ``Analysis of the effect of
  low-overhead lossy image compression on the performance of visual crowd
  counting for smart city applications,'' \emph{ISC2}, 2022.

\bibitem{zhang2016single}
Y.~Zhang, D.~Zhou \emph{et~al.}, ``Single-image crowd counting via multi-column
  convolutional neural network,'' \emph{CVPR}, 2016.

\bibitem{hudson2017jpeg}
G.~Hudson, A.~L{\'e}ger \emph{et~al.}, ``Jpeg at 25: Still going strong,''
  \emph{IEEE MultiMedia}, 2017.

\bibitem{gueguen2018faster}
L.~Gueguen, A.~Sergeev \emph{et~al.}, ``Faster neural networks straight from
  jpeg,'' \emph{NeurIPS}, 2018.

\bibitem{dodge2016understanding}
S.~Dodge and L.~Karam, ``Understanding how image quality affects deep neural
  networks,'' \emph{QoMEX}, 2016.

\bibitem{ehrlich2021analyzing}
M.~Ehrlich, L.~Davis \emph{et~al.}, ``Analyzing and mitigating jpeg compression
  defects in deep learning,'' \emph{ICCV}, 2021.

\bibitem{jiang2021towards}
J.~Jiang, K.~Zhang, and R.~Timofte, ``Towards flexible blind jpeg artifacts
  removal,'' \emph{CVPR}, 2021.

\bibitem{gao2020cnn}
G.~Gao, J.~Gao \emph{et~al.}, ``Cnn-based density estimation and crowd
  counting: A survey,'' \emph{arXiv}, 2020.

\bibitem{10.1145/3460426.3463628}
F.~Dai, H.~Liu \emph{et~al.}, ``Dense scale network for crowd counting,''
  \emph{ICMR}, 2021.

\bibitem{bengio2009curriculum}
Y.~Bengio, J.~Louradour \emph{et~al.}, ``Curriculum learning,'' \emph{ICML},
  2009.

\bibitem{DBLP:journals/corr/abs-2010-13166}
X.~Wang, Y.~Chen, and W.~Zhu, ``A comprehensive survey on curriculum
  learning,'' \emph{CoRR}, 2020.

\bibitem{pmlr-v97-hacohen19a}
G.~Hacohen and D.~Weinshall, ``On the power of curriculum learning in training
  deep networks,'' \emph{ICML}, 2019.

\bibitem{bakhtiarnia2021improving}
A.~Bakhtiarnia, Q.~Zhang, and A.~Iosifidis, ``Improving the accuracy of early
  exits in multi-exit architectures via curriculum learning,'' \emph{IJCNN},
  2021.

\bibitem{9275392}
Q.~Wang, W.~Lin \emph{et~al.}, ``Density-aware curriculum learning for crowd
  counting,'' \emph{IEEE Transactions on Cybernetics}, 2022.

\bibitem{hu2020ambient}
D.~Hu, L.~Mou \emph{et~al.}, ``Ambient sound helps: Audiovisual crowd counting
  in extreme conditions,'' \emph{arXiv}, 2020.

\bibitem{song2021choose}
Q.~Song, C.~Wang \emph{et~al.}, ``To choose or to fuse? scale selection for
  crowd counting,'' \emph{AAAI}, 2021.

\bibitem{simonyan2014very}
K.~Simonyan and A.~Zisserman, ``Very deep convolutional networks for
  large-scale image recognition,'' \emph{ICLR}, 2015.

\bibitem{li2018csrnet}
Y.~Li, X.~Zhang, and D.~Chen, ``Csrnet: Dilated convolutional neural networks
  for understanding the highly congested scenes,'' \emph{CVPR}, 2018.

\bibitem{paszke2019pytorch}
A.~Paszke, S.~Gross \emph{et~al.}, ``Pytorch: An imperative style,
  high-performance deep learning library,'' \emph{NeurIPS}, 2019.

\bibitem{loshchilov2018decoupled}
I.~Loshchilov and F.~Hutter, ``Decoupled weight decay regularization,''
  \emph{ICLR}, 2019.

\bibitem{yang2021compression}
E.-H. Yang, H.~Amer, and Y.~Jiang, ``Compression helps deep learning in image
  classification,'' \emph{Entropy}, 2021.

\end{thebibliography}

\end{document}